\documentclass[10pt,letterpaper]{article}
\usepackage[top=0.85in,left=2.75in,footskip=0.75in,marginparwidth=2in]{geometry}

\usepackage[utf8]{inputenc}

\usepackage{cite}

\usepackage{algorithm2e}

\usepackage{nameref,hyperref}

\usepackage[right]{lineno}

\usepackage{microtype}
\DisableLigatures[f]{encoding = *, family = * }

\raggedright
\usepackage{changepage}

\usepackage[aboveskip=1pt,labelfont=bf,labelsep=period,singlelinecheck=off]{caption}

\makeatletter
\renewcommand{\@biblabel}[1]{\quad#1.}
\makeatother

\usepackage{lastpage,fancyhdr,graphicx}
\usepackage{epstopdf}
\fancyheadoffset[L]{2.25in}
\fancyfootoffset[L]{2.25in}

\usepackage{color}

\definecolor{Gray}{gray}{.25}

\usepackage{graphicx}
\usepackage{multirow}
\usepackage{amsmath}
\usepackage{amssymb}
\usepackage{algorithmic}
\usepackage{subfig}

\usepackage{sidecap}

\usepackage{wrapfig}
\usepackage[pscoord]{eso-pic}
\usepackage[fulladjust]{marginnote}
\reversemarginpar

\begin{document}
\vspace*{0.35in}


\begin{flushleft}
{\Large
\textbf\newline{Distributed Averaging CNN-ELM for Big Data}
}
\newline
\\
Arif Budiman,
Mohamad Ivan Fanany,
Chan Basaruddin,
\\
\bigskip
\bf{1} Machine Learning and Computer Vision Laboratory \\Faculty of Computer Science, Universitas Indonesia\\
\bigskip
* intanurma@gmail.com

\end{flushleft}

\providecommand{\keywords}[1]{\textbf{\textit{Keywords---}} #1}

\section*{Abstract}
Increasing the scalability of machine learning to handle big volume of data is a challenging task. The scale up approach has some limitations. In this paper, we proposed a scale out approach for CNN-ELM based on MapReduce on classifier level. Map process is the CNN-ELM training for certain partition of data. It involves many CNN-ELM models that can be trained asynchronously. Reduce process is the averaging of all CNN-ELM weights as final training result. This approach can save a lot of training time than single CNN-ELM models trained alone. This approach also increased the scalability of machine learning by combining scale out and scale up approaches. We verified our method in extended  MNIST data set and not-MNIST data set experiment. However, it has some drawbacks by additional iteration learning parameters that need to be carefully taken and training data distribution that need to be carefully selected. Further researches to use more complex image data set are required. 
\bigskip

\noindent\keywords{deep learning, extreme learning machine, convolutional, neural network, big data, map reduce}


\section{Introduction}

Nowadays, We are seeing a massive growth of data at a faster rate than ever before. However, the benefits of big data  become meaningless if none of the processing machine can cultivate and adapt to the data quickly enough. Big data mining needs special machine learning approaches to learn huge volumes of data in an acceptable time. Volume and velocity issues are critical in overcoming big data challenges \cite{laney:1}. It means the data are so massive hence very difficult to be handled by a single computation task in a timely fashion. As with many new hardware and software technologies, we require a special approach to make the most of hardware and software work effectively on speed, scalability, and simplicity presents in real big data knowledge mining. 

Scalability is the ability of data processing  system to adapt against increased demands. It can be categorized into the following two types of scalability \cite{Singh2014}:
\begin{enumerate}
\item Vertical Scaling: Known as scale up. It involves powering more and larger computation components within a single system. It is also known as "scale up" and it usually involves a single instance of an operating system. I.e. Adding more power and capacity (CPU, GPU, RAM, Storage) to an existing machine. However, Scale up is limited to the maximum hardware specification of a single machine.
\item Horizontal Scaling: Known as scale out. The system  distributes the workload across many independent computation resources which may be low end commodity machines or high end machines. All resources added together can speed up the processing capability. Thus we add more machines into one pool of resources. Scale out offers easier and dynamic scalability by adding various size of machines into the existing pool. 

\end{enumerate}  

To increase the scalability of big data processing, the common approach is to distribute big data and running the process in parallel. Parallel computing is a simultaneous use of multiple computing resources to solve complex computational problems by breaking down the process into simpler series of instructions that can be executed simultaneously on different processing units and then employ an overall control management \cite{barney:1}. 

To overcome overhead complexities in parallel programming, Google introduced a programming model named MapReduce \cite{Dean:2008}. 
MapReduce is a framework for processing large data  within a parallel and distributed way on many  computers including low end computers in a cluster.  
MapReduce provides two essential functions: 1) Map function, it processes each sub problems to another nodes within the cluster; 2) Reduce function, it organizes the results from each node to be a cohesive solution \cite{10.1109/FCCM.2008.19}. 

Developing MapReduce is simple by firstly exposing structure and process similarity and then aggregation process \cite{10.1109/FCCM.2008.19}. All the similar tasks are easily parallelized, distributed to the processors and load balanced between them. MapReduce framework does not related to specific hardware technologies. It can be employed to multiple and heterogeneous machine independent.

Further researches introduced MapReduce paradigm to speed up various machine learning algorithms, i.e., locally weighted linear regression (LWLR), k-means, logistic regression
(LR), naive Bayes (NB), support vector machine (SVM), gaussian discriminant analysis (GDA), expectation–maximization (EM) and backpropagation (NN) \cite{NIPS2006_3150}, stochastic gradient descent (SGD) \cite{NIPS2010_4006}, convolutional neural network (CNN) \cite{Wang2016}, extreme learning machine (ELM) \cite{Xin2015464}. 

CNN \cite{LeCun:1} is a popular machine learning that getting benefits from parallel computation. CNN uses a lot of convolution operations that needs many processing cores to speed up the computation time using  graphics processing units (GPUs) parallelization. However, the scale up approach still has limitation mainly caused by the amount of memory available on GPUs \cite{NIPS2012_4824,Scherer2010}.   

Learn from the scale up limited capability, we proposed a scale out approach based on MapReduce model to distribute the big data computation into several CNN models. We integrated the CNN architecture  \cite{LeCun:1,Zeiler2014,Jiuxiang:1} with  ELM \cite{huang2006extreme,Huang:ELMSurvey,Huang201532,Xin2015464}. The CNN works as  unsupervised convolution features learner and  ELM works as supervised classifier. We employed parallel stochastic gradient descent (SGD) algorithm \cite{NIPS2010_4006} to fine tune the weights of CNN-ELM and to average the final weights of CNN-ELM. 

Our main contributions in this paper are as follows.
\begin{enumerate}
    \item We studied the CNN-ELM integration using MapReduce model;
    \item We employed map processes as  CNN-ELM multi classifiers learning independently (asynchronous) on different partition of training data. The reduce process is the averaging all weights (kernel weights on CNN and output weights on ELM) of all CNN-ELM classifiers. Our method enables scale out combination of highly scale up CNN-ELM members to handle very huge training data. Our idea is to place  MapReduce model not intended for CNN matrix operation level but for classifier level. Many asynchronous CNN models trained together to solve very large complex problem rather than single models trained in very powerful machine.  
    \item  Against ELM tenet for non iterative training, we studied the  weight after fine tuning using stochastic gradient descent iteration during ELM training to check the averaging performance after some iterations. 
\end{enumerate}

The rest of this paper is organized as follows. Section 1 is to give introduction and research objectives. In Section 2, a related review of previous MapReduce framework implementations is given. Section 3 is to describe our proposed methods. Our empirical experiments result is introduced in Section 4. Finally, conclusions are drawn in Section 5.

\section{Literature Reviews}

\subsection{Parallel SGD and weight averaging}

SGD is a very popular training algorithm for various machine learning models i.e.,  regression, SVM, and NN.  Zinkevich \textit{et.al} \cite{NIPS2012_4824} proposed a parallel model of  SGD based on MapReduce that highly suitable for parallel and  large-scale machine learning. In parallel SGD, the training data is accessed locally by each model and only communicated when it finished. The algorithm of parallel SGD is described below \ref{simuSGD}.

\newcommand\ParDo{\renewcommand\algorithmicdo{\textbf{parallel do}}}
\newcommand\noParDo{\renewcommand\algorithmicdo{\textbf{do}}}

\begin{algorithm}
\caption{SimuParallelSGD(Training \{$x^1$,...,$x^m$\}; Learning Rate $\eta$; Machines $k$)}\cite{NIPS2012_4824} 
\label{simuSGD}
\begin{algorithmic}[1]
\STATE Define $ P=\lfloor m/k \rfloor$ \\
\STATE Randomly partition the training, giving $P$ examples to each machine.
\ParDo
\FORALL{$i\in \{1,...,k\}$} 
    \STATE Randomly shuffle the data on machine $i$
    \STATE Initialize $w_{i,0}=0$
        \noParDo
        \FORALL{$p\in \{1,...,P\}$} 
        \STATE Get the $p^{th}$ training on the $i^{th}$ machine $c^{i,p}$
        \STATE $w_{i,p} \leftarrow w_{i,p-1} - \eta \delta c^i (w_{i,p-1})$
        \ENDFOR
\ENDFOR
\STATE Aggregate from all machines $v= \frac{1}{k}\sum_{i=1}^{k} w_i$
\RETURN $v$
\end{algorithmic}
\end{algorithm}

The idea of averaging was developed by Polyak \textit{et.al}  \cite{doi:10.1137/0330046}. The averaged SGD is ordinary SGD that averages  its  weight over time. When optimization is finished, the averaged weight replaces the ordinary weight from SGD. It is based on the idea of averaging the trajectories, however the application requires a large amount of a priori information. 

Let we have unlimited training data : $\left \{(\mathbf{x_{(0)}},\mathbf{t_{(0)}}),(\mathbf{x_{(1)}},\mathbf{t_{(1)}}),\cdots,(\mathbf{x_{(\infty)}},\mathbf{t_{(\infty)}})\right\}$ within the same distribution. 
Learning objective is to construct the mapping function $\hat{\boldsymbol{\beta}}$ from observation data that taken randomly and its related class. However, when the number of training data  $\textstyle{m}\rightarrow\infty$, we need to address the expected value of $ \hat{\boldsymbol{\beta}}(w)$ with $w$ is the learning parameters. According to law of large numbers, we can make sure the consistency of expected value of learning model is $\hat{\boldsymbol{\beta}}(w) $ approximated by the sample averages $\frac{1}{m}\sum_{i=1}^{m} \hat{\boldsymbol{\beta}}(w)_i $ and almost surely to the expected value as $\textstyle{m}\rightarrow\infty$ with probability 1.

If the $m$ training data is partitioned by $k$ to be $T$ partition, and each partition trained independently  $\left\{  \hat{\boldsymbol{\beta}}(w)_0,\hat{\boldsymbol{\beta}}(w)_1,...,\hat{\boldsymbol{\beta}}(w)_T\right\}$, we can make sure the expected value $\hat{\boldsymbol{\beta}}(w) $ is approximated by $\frac{1}{T}\sum_{i=1}^{T} \hat{\boldsymbol{\beta}}(w)_i $ where $ T=\lfloor m/k \rfloor$. 

\subsection{MapReduce in ELM}

Extreme Learning Machine (ELM) is one of the famous machine learning that firstly proposed by Huang \cite{huang2006extreme,Huang:ELMSurvey,Huang201532}. It used single hidden layer feedforward neural network (SLFN) architecture and generalized pseudoinverse for learning process. Similar with Neural Networks (NN), ELM used random value in hidden nodes parameters. The uniqueness of ELM is  non iterative generalized pseudoinverse optimization process However, the hidden nodes parameters remain set and fixed after the training. It becomes the ELM training is fast and can avoid local minima. 

The ELM learning result  is Output weight ($\beta$) that can be computed by:

\begin{equation}
\label{eq:2}
\hat{\beta} = \textbf{H}^{\dagger}\textbf{T}
\end{equation}

which $\textbf{H}^{\dagger }$ is a pseudoinverse (Moore-Penrose generalized inverse) function of $\textbf{H}$. The ELM learning objective is to find the smallest least-squares solution  of linear system $\textbf{H}\beta - \textbf{Y}$ that can be obtained when $\hat{\beta}$ = $\textbf{H}^{\dagger}\textbf{T}$. 

Hidden layer matrix $\textbf{H}$ is computed by activation function $\textit{g}$ of the summation matrix  from the hidden nodes parameter (such as input weight $a$ and bias  $b$) and training input \textbf{x} with size N number of training data and L number of hidden nodes $g(a_{i}\cdot\textbf{x}+b_{i})$ (called random feature mapping). 

The performance of ELM hinges on generalized inverse solution. The solution of $\textbf{H}^{\dagger}$ uses ridge regression orthogonal projection  method, by using a positive  1/$\lambda$ value as regularization to the auto correlation matrices  $\textbf{H}^{T}\textbf{H}$ or $\textbf{H}\textbf{H}^{T}$. Thus, we can solve Eq. \ref{eq:2} as follows. 

\begin{equation}
	\label{eq:orthogonalization}
	\beta = \left( \frac{\textbf{I}}{\lambda}+\textbf{H}^{T}\textbf{H}\right)^{-1}\textbf{H}^{T}\textbf{T}
\end{equation}

Further, Eq. \ref{eq:orthogonalization} can be solved by sequential series using block matrices inverse (A Fast and Accurate Online Sequential named online sequential extreme learning machine (OS-ELM) \cite{4012031}) or by MapReduce approach (Elastic Extreme Learning Machine (E$^{2}$LM) \cite{Xin2015464}  or Parallel ELM \cite{He:2013}).

Parallelization process using MapReduce approach can be divided as follows :
\begin{enumerate}
\item Map.
Map is the transformation of intermediate matrix multiplications for each training data and target portion.  

    If $\mathbf{U}= \mathbf{H}^{T}\mathbf{H}$ and $\mathbf{V}= \mathbf{H}^{T}\mathbf{T}$, According to decomposable matrices, they can be written as : 
    
    \begin{equation}
	\label{eq:decomU}
	\mathbf{U} = \sum_{k=0}^{k=\infty} \mathbf{U}_{(k)}
    \end{equation}

    \begin{equation}
	\label{eq:decomV}
	\mathbf{V} = \sum_{k=0}^{k=\infty} \mathbf{V}_{(k)}
    \end{equation}

\item Reduce.
Reduce is the aggregate process to sum the Map result.
The output weights $\beta$ can be computed easily from reduce/aggregate process. 

    \begin{equation}
	\label{eq:elasticELM}
	\beta = \left( \frac{\textbf{I}}{\lambda}+\textbf{U}\right)^{-1}\textbf{V}
    \end{equation}
\end{enumerate}

Therefore, MapReduce based ELM  is more efficient for massive training data set, can be solved easily by parallel computation and has better performance  \cite{Xin2015464}. 

Regarding about iteration, Lee \textit{et.al} \cite{cnn:lee} explained on BP Trained Weight-based ELM that the optimized input weights with BP training is more feasible than randomly assigned weights. Lee \textit{et.al} implemented Average ELM however the classification accuracy was lower than basic ELM because the number of training data is so small and the network architecture is not large.

\subsection{MapReduce in CNN}

CNN is biologically-inspired \cite{LeCun:1} from visual cortex that has a convolution arrangement of cells (a receptive field that sensitive to small sub visual field and local filter) and following by simple cells that only respond maximally to specific trigger within receptive fields. A simple CNN architecture consists of some convolution layers and following by pooling layers in the feed forward architecture. CNN has excellent performance for spatial visual classification \cite{Simard:2003}. 

The input layer exposes 2D structure with $d \times d \times r$ of image, and $r$ is the number of input channels. The convolution layer has $c$ filters (or kernels) of size $k \times k \times q$  where $k < d$  and $q$ can either be the same or smaller than the number of input channels $r$. The filters have locally connected structure which is each convolved with the image to produce $c$ feature maps of size $d - k+1$. If, at a given layer, we have the $r^{th}$ feature map  as $h^r$, whose filters are determined by the weights $W^r$ and bias $b_r$, then the feature map $h^r$ is obtained as :

\begin{equation}
	\label{eq:conv}
	h^{r}_{ij} = g((W^r \ast x)_{ij} + b_r)
\end{equation}

\begin{figure}[!h]
\centering
\includegraphics[width=5in]{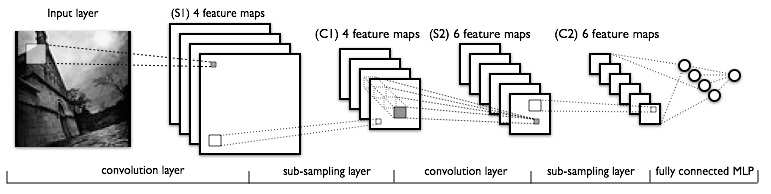}
\caption{CNN Architecture from LeNet}. 
\label{fig:cnnarch}
\end{figure}

Each feature map is then pooled using pooling operation either down sampling, mean or max sampling over $s \times s \times s$ contiguous regions (Using scale $s$ ranges between 2 for small and up to 5 for larger inputs). 
An additive bias and activation function (i.e. sigmoid, tanh, or reLU) can be applied to each feature map either before or after the pooling layer. At the end of the CNN layer, there may be any densely connected NN layers for supervised learning (See Fig. \ref{fig:cnnarch}) \cite{Zeiler2014}. Many variants of CNN architectures in the literatures, but the basic common building blocks are convolutional layer, pooling layer and fully connected layer \cite{Jiuxiang:1}.

The convolution operations need to be computed in parallel for faster computation time that can be taken from multi processor hardware, i.e., GPU \cite{Scherer2010}. Krizhevsky \textit{et. al.}  demonstrated a large CNN is capable of achieving record breaking results on the 1.2 million high-resolution images with 1000 different classes. However, the GPU has memory size limitation that limit the CNN network size to achieve better accuracy \cite{NIPS2012_4824}.

CNN used back propagation algorithm that needs iterations to get the optimum solution. One iteration contains error back propagated step and following by parameter update step. The learning errors are propagated back to the previous layers using SGD optimization  and continued by applying the  update to kernel weight  and bias parameters. 

If $\delta^{(l+1)}$ is the error on $(l+1)^{th}$ layer from a cost function $J(W,b;x,t)$  where $W$ is weight, $b$ is bias parameters, and $(x,t)$ are the training data and target. 

\begin{equation}
	\label{eq:J}
	J(W,b;x,t) = \frac{1}{2}\parallel f(z) -t \parallel^2
\end{equation}

If the $l^{th}$ layer  is densely connected and the $(l+1)^{th}$ is output layer, then the error  $\delta^{(l)}$ and the gradients for the $l^{th}$ layer are computed as :

\begin{equation}
	\label{eq:conv1}
	\delta^{(l)} = \big( (W^{(l)})^T\delta^{(l+1)} \big) \cdot f'(z^{(l)})
\end{equation}
Where $f'$ is the derivative of the activation function.

\begin{gather} 
\label{eq:grad}
\nabla_{W^{(l)}}J(W,b;x,t) = \delta^{(l+1)}(a^{(l)})^T \\ 
\nabla_{b^{(l)}}J(W,b;x,t) = \delta^{(l+1)}
\end{gather}

But if the $l^{th}$ layer is a convolutional and subsampling layer then the error is computed as :

\begin{equation}
	\label{eq:conv2}
	\delta_{r}^{(l)} = pool \big( (W_{r}^{(l)})^T\delta_{r}^{(l+1)} \big) \cdot f'(z_{r}^{(l)})
\end{equation}
Where $pool$ is the related pooling operation.  

To calculate the gradient to the filter maps, we used convolution operation and flip operation to the error matrix. 

\begin{gather} 
\label{eq:grad1}
\nabla_{W_{r}^{(l)}}J(W,b;x,t) = \sum_{i=1}^m (a_i^{(l)}) \ast rot90 (\delta_{r}^{(l+1)},2) \\ 
\nabla_{b_{r}^{(l)}}J(W,b;x,t) = \sum_{a,b}(\delta^{(l+1)})_{a,b}
\end{gather}
Where $a^{(l)}$ is the input of $l^{th}$ layer, and $l=1$ is the input layer.

Finally, one iteration will  update the parameters $W$ and $b$ with $\alpha$ learning rate,  as follows:

\begin{gather} 
\label{eq:grad2}
W_{ij}^{(l)} = W_{ij}^{(l)} - \alpha\frac{\partial}{\partial W_{ij}^{(l)}}J(W,b) \\ 
b_{i}^{(l)} = b_{i}^{(l)} - \alpha\frac{\partial}{\partial b_{i}^{(l)}}J(W,b)
\end{gather}

Most CNN implementations are using GPU \cite{Scherer2010} to speed up convolution operation that required hundred numbers of core processors. Wang \textit{et. al} \cite{Wang2016} used MapReduce on  Hadoop platform to take advantage of the computing power of multi core CPU to solve matrix parallel computation. However, the number of  multi core CPU is far less than GPU can provide. 

GPU has limited shared memory than CPU global memory. Scherer \textit{et. al} \cite{Scherer2010} explained because shared memory is very limited, so it reuses loaded data as often as possible. Comparing with CPU, the global memory in CPU can be extended larger with lower price than additional GPU cards.

\section{Proposed Method}

\begin{figure}[!h]
\centering
\includegraphics[width=3.5in]{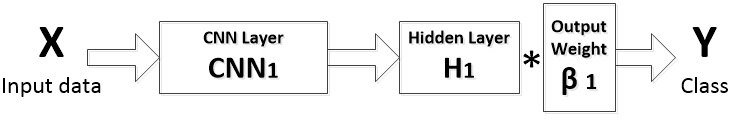}
\caption{CNN-ELM integration architecture : The last convolution layer output is submitted as hidden nodes weight \textbf{H} of ELM}. 
\label{fig:method2}
\end{figure}

We used common CNN-ELM integration \cite{ShanPang_hindawi,bai2015,cnn:guo} architecture   when the last convolution layer output is fed as hidden nodes weight \textbf{H} of ELM (See Fig. \ref{fig:cnnarch}). For better generalization accuracy, we used nonlinear  optimal tanh ($1.7159\times tanh (\frac{2}{3}\times\mathbf{H}$) activation function \cite{LeCun:1998}. 
We used the E$^2$LM as a parallel supervised classifier to replace fully connected NN. Compared with regular ELM, we do not need input weight as hidden nodes parameter (See Fig. \ref{fig:method2}).


\begin{figure}[!h]
    \centering
    \includegraphics[width=3.5in]{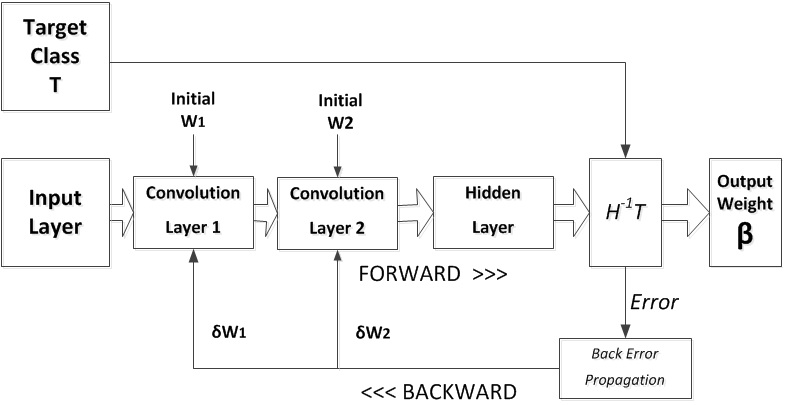}
    \caption{CNN Architecture}. 
    \label{fig:method1}
    \end{figure}

The idea of backward is similar with densely connected NN back propagation error method with cost function : 
\begin{equation}
	\label{eq:JELM}
	J(\beta;z,t) = \frac{1}{2}\parallel \mathbf{H}(z)\beta -\mathbf{T} \parallel^2
\end{equation}
Then  it propagated back with SGD to optimize the weight kernels of convolution layers (See Fig. \ref{fig:method1}). 

Detail algorithm is explained on Algorithm \ref{simuCNNELM}.
\begin{algorithm}
\caption{Distributed CNNELM(Training \{$x^1$,...,$x^m$\}; Learning Rate $\eta$; Machines $k$; iterations $e$)}
\label{simuCNNELM}
\begin{algorithmic}[1]
\STATE Define $ P=\lfloor m/k \rfloor$ \\
\STATE Randomly partition the training, giving $P$ examples to each machine.
\STATE Initialize CNN weight parameters similar for $k$ machines
\ParDo
\FORALL{$i\in \{1,...,k\}$} 
    \STATE Randomly shuffle the data on machine $i$
    \noParDo
    \FORALL{$j\in \{1,...,e\}$}
        \STATE Reset $\Sigma U = 0$;$\Sigma V = 0$
        \FORALL{$p\in \{1,...,P\}$} 
        \STATE Get \textbf{H} from  $p^{th}$ CNN training on the $i^{th}$ machine $c^{i,p}$
        \STATE Compute $\Sigma U = \Sigma U + H^TH$
        \STATE Compute $\Sigma V = \Sigma U + H^TT$
        \STATE Compute $\beta$
        \STATE Propagate ELM Error back to CNN
        \STATE Update Kernel Weights $W_{ij}^{(l)}$ and bias $b_{ij}^{(l)}$
        \ENDFOR
    \ENDFOR
\ENDFOR
\STATE Aggregate for each $l^{th}$ layers $\hat{W}^{(l)}= \frac{1}{k}\sum_{i=1}^{k} W_i^{(l)}$
\STATE Aggregate for each $l^{th}$ layers $\hat{b}^{(l)}= \frac{1}{k}\sum_{i=1}^{k} b_i^{(l)}$
\STATE Aggregate for each $l^{th}$ layers $\hat{\beta} = \frac{1}{k}\sum_{i=1}^{k} \beta_i$
\RETURN $\hat{W}$, $\hat{b}$, $\hat{\beta}$
\end{algorithmic}
\end{algorithm}

\section{Experiment and Performance Results}

\subsection{Data set}

\begin{figure}[!h]
\centering
\includegraphics[width=2in]{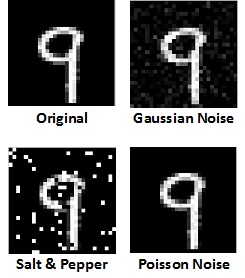}
\caption{Extended MNIST Data set by adding random gaussian, salt\&pepper, poisson noise to original data. }. 
\label{fig:noise}
\end{figure}

MNIST is the common data set for big data machine learning, in fact, it accepted as standard and give an excellent result. MNIST data set is a balanced data set that contains numeric (0-9) (10 target class) with size $28 \times 28$ pixel in a gray scale image. The dataset has been divided for 60,000 examples for training data and separated 10,000 examples for testing data  \cite{lecun-mnisthandwrittendigit-2010}. We  extended MNIST data set $3 \times$ larger by adding 3 types of image noises (See Fig. \ref{fig:noise}) to be 240,000 examples of training data and 40,000 examples of testing data.

\begin{figure}[!h]
\centering
\includegraphics[width=2in]{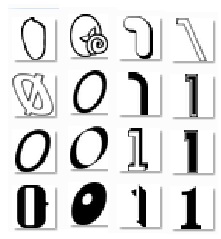}
\caption{Not-MNIST Data set for numeric (0-9) symbols }. 
\label{fig:nonMNIST1}
\end{figure}

\begin{figure}[!h]
\centering
\includegraphics[width=2in]{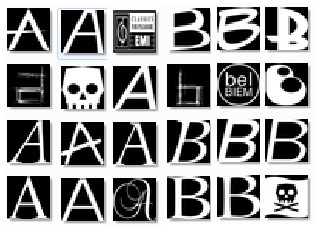}
\caption{Not-MNIST Data set for alphabet (A-J) symbols }. 
\label{fig:nonMNIST}
\end{figure}

For additional experiments, we used not-MNIST large data set \cite{notMNIST1} that has a lot of foolish images (See Fig. \ref{fig:nonMNIST1} and \ref{fig:nonMNIST}).  Not-MNIST has gray scale $28 \times 28$ image size as attributes. We divided the set to be numeric (0-9) (360,000 data) and alphabet (A-J) symbol (540,000) data including many foolish images. The challenge with not-MNIST numeric and not-MNIST alphabet is many similarities between class 1 with class I, class 4 with class A, and another look alike foolish images. 

\subsection{Experiment Methods}

We defined the scope of works as following:
\begin{enumerate}
 
\item We enhanced DeepLearn Toolbox \cite{DLtoolbox} with Matlab parallel computing toolbox. 
\item We used single precision  for all computation in this paper. 
\item We focused on simple CNN architecture that consist of convolution layers (c), following by \textit{reLU} activation layer  then pooling layer (s) with down sampling  in this paper. 
\item We compared the performance in testing accuracy with non partitioned sequential CNN-ELM classifier using the similar structure size. 

\end{enumerate}

\begin{table}[!h]

\renewcommand{\arraystretch}{1.3}
    \caption{Data set Dimension, Quantity, and Evaluation method.}
    \label{alltable2}
    \centering
    \subfloat[Data Set dimension and  Quantity\label{table:dataset}]{%
\centering     
\begin{tabular}{|c|c|c|c|}
\hline
Data Set Concepts &  Inputs & Outputs & Data   \\
\hline
\hline
$\mathbf{MNIST}$ & 784  & 10 (0-9)    & 240,000   \\
\hline
$\mathbf{Not MNIST}$& 784  & 20 (0-9,A-J)     & 900,000   \\
\hline
\end{tabular}
    }
    \\
    \subfloat[Evaluation  Method\label{table:eval}]{%
\centering     
\begin{tabular}{|p{1.25cm}|p{2cm}|c|c|}
\hline
Data Set & Evaluation Method & Training & Testing \\
\hline
\hline
MNIST &  Holdout ($5\times$ trials on different computers)   & 240,000 & 40,000 \\
\hline
Not-MNIST &  Cross Validation 6 Fold   & 750,000 & 150,000 \\
\hline
\end{tabular}
}
     \\
     \subfloat[Performance Measurements\label{table:mea}]{%
 \centering     
 \begin{tabular}{|p{2cm}|p{4cm}|}
 \hline
 Measure & Specification \\
 \hline
 \hline
 Accuracy &  The accuracy of classification in \% from $\frac{\#\textit{Correctly Classified}}{\#\textit{Total Instances}} $ \\
 \hline
 Testing Accuracy &  The accuracy measurement of the testing data which not part of training set. \\
 \hline
 Cohen's Kappa and kappa error  &  The  statistic measurement of inter-rater agreement for categorical items. \\
 \hline
 \end{tabular}
     }
  \end{table}

To verify our method, we formulated the following research questions: 
\begin{itemize}

\item How is the performance following number of iterations?
\item How is the effectiveness of weight averaging CNN-ELM model for various number of training partition? 
\item How is the performance consistencies of weight averaging CNN-ELM model following number of iterations?

\end{itemize}

\subsection{Performance Results}


In this section, we explained the research questions as follows.

\begin{itemize}

        \item The performance of CNN-ELM can be improved by using back propagation algorithm. However, we need to select the appropriate learning rate parameter, number of batch and number of  iteration that could impact to the final performance (See Fig. \ref{fig:picture_epoch}). The wrong parameter selection especially learning rate could trap into local minima. So, we can use dynamic learning rate rather than static rate. 

\begin{figure}[!h]
\centering
\begin{tabular}{c}
\subfloat[Testing Accuracy for 3c-2s-6c-2s kernel size=5 CNN-ELM Model ]{\includegraphics[width=3in]{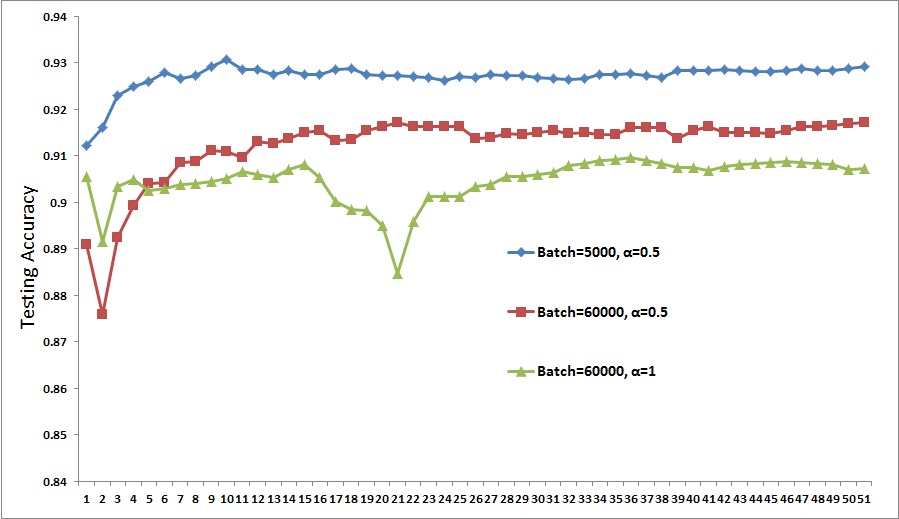}} \\ 
\subfloat[Performance dropped because of wrong learning rate $\alpha$ parameter ]{\includegraphics[width=3in]{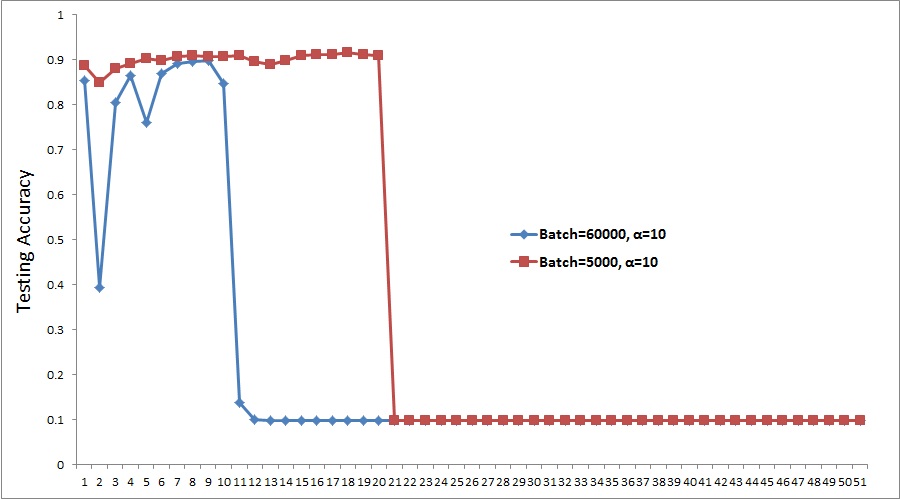}} 
\end{tabular}
\caption{Testing Accuracy on extended MNIST data set using 6c-2s-12c-2s  CNN-ELM Model. }
\label{fig:picture_epoch}
\end{figure}

        \item In this experiment, we partitioned the training data to be 2 partitions and 5 partitions on not-MNIST. We compared the testing accuracy of CNN-ELM no partition model with average 2 partition model and average 5 partition model (See table \ref{not-MNIST0} and \ref{not-MNIST1}). Unfortunately, the performance of average CNN-ELM more partitions and more iterations model has decreased than CNN-ELM no partition model. However, different result found for extended MNIST (See table \ref{MNIST0} and \ref{MNIST1}). Because extended MNIST  has been built from the same distribution on each 60,000 partition size while not on not-MNIST. 
        
\begin{table}[!h]

\renewcommand{\arraystretch}{1.3}
    \caption{Testing Accuracy for 3c-2s-9c-2s kernel size=5 at iteration=0, batch=75,000 on not-MNIST}
    \label{not-MNIST0}
    \centering
    \begin{tabular}{|c|c|c|}
\hline
Model &   Testing Accuracy \%  \\
\hline
\hline
CNN-ELM 1   & 72.85$\pm$1.23       \\
\hline
CNN-ELM 1/2   & 40.51$\pm$0.87       \\
\hline
CNN-ELM 2/2   & 40.35$\pm$0.86         \\
\hline
CNN-ELM Average 2 & 67.91$\pm$2.77       \\
\hline
CNN-ELM 1/5   & 20.56$\pm$0.22       \\
\hline
CNN-ELM 2/5   & 20.21$\pm$0.94        \\
\hline
CNN-ELM 3/5   & 20.50$\pm$0.91        \\
\hline
CNN-ELM 4/5   & 31.48$\pm$0.54         \\
\hline
CNN-ELM 5/5   & 31.47$\pm$0.53         \\
\hline
CNN-ELM Average 5 & 60.83$\pm$0.20       \\
\hline
\end{tabular}
    
\end{table}    

\begin{table}[!h]

\renewcommand{\arraystretch}{1.3}
    \caption{Testing Accuracy for 3c-2s-9c-2s kernel size=5 at iteration e=5, $\alpha=\frac{5}{e}$ , batch=75,000 on not-MNIST}
    \label{not-MNIST1}
    \centering
    \begin{tabular}{|c|c|c|}
\hline
Model &   Testing Accuracy \%  \\
\hline
\hline
CNN-ELM 1   & 73.72$\pm$1.32       \\
\hline
CNN-ELM 1/2   & 41.45$\pm$1.25       \\
\hline
CNN-ELM 2/2   & 41.19$\pm$0.73         \\
\hline
CNN-ELM Average 2 & 66.85$\pm$2.43       \\
\hline
CNN-ELM 1/5   & 20.56$\pm$0.24       \\
\hline
CNN-ELM 2/5   & 20.09$\pm$0.96       \\
\hline
CNN-ELM 3/5   & 21.22$\pm$0.86        \\
\hline
CNN-ELM 4/5   & 31.71$\pm$0.52         \\
\hline
CNN-ELM 5/5   & 31.70$\pm$0.52         \\
\hline
CNN-ELM Average 5 &  59.59$\pm$0.24       \\
\hline
\end{tabular}
    
\end{table} 

\begin{table}[!h]

\renewcommand{\arraystretch}{1.3}
    \caption{Testing Accuracy for 6c-2s-12c-2s kernel size=5 at iteration=0, batch=60,000 on MNIST}
    \label{MNIST0}
    \centering
    \begin{tabular}{|c|c|c|}
\hline
Model &   Testing Accuracy \%  \\
\hline
\hline
CNN-ELM 1   & 92.23$\pm$0.44       \\
\hline
CNN-ELM 1/4   & 92.13$\pm$0.87       \\
\hline
CNN-ELM 2/4   & 92.22$\pm$0.43         \\
\hline
CNN-ELM 3/4   & 92.16$\pm$0.23       \\
\hline
CNN-ELM 4/4   & 92.11$\pm$0.13        \\
\hline
CNN-ELM Average 4 & 92.24$\pm$0.23       \\
\hline
\end{tabular}

\end{table}   

\begin{table}[!h]

\renewcommand{\arraystretch}{1.3}
    \caption{Testing Accuracy for 6c-2s-12c-2s kernel size=5 at iteration e =5, $\alpha=\frac{1}{e}$, batch=60,000 on MNIST}
    \label{MNIST1}
    \centering
    \begin{tabular}{|c|c|c|}
\hline
Model &   Testing Accuracy \%  \\
\hline
\hline
CNN-ELM 1   & 92.41$\pm$0.36       \\
\hline
CNN-ELM 1/4   & 92.26$\pm$0.13       \\
\hline
CNN-ELM 2/4   & 92.37$\pm$0.56         \\
\hline
CNN-ELM 3/4   & 92.20$\pm$0.31       \\
\hline
CNN-ELM 4/4   & 92.28$\pm$0.17       \\
\hline
CNN-ELM Average 4 & 92.40$\pm$0.26       \\
\hline
\end{tabular}
\end{table}
        
\end{itemize}

\section{Conclusion}

The proposed CNN-ELM method gives better scale out capability for processing large data set in parallel. We can partition large data set. We can assign CNN-ELM classifier for each partition, then we just aggregated the result by averaging the weight parameters of all CNN-ELM parameters. Thus, it can safe a lot of training time rather than sequential training. However, more CNN-ELM classifiers (smaller partition) has worse performance for averaging CNN-ELM, as well as more iterations  and data distribution effect.

We think some ideas for future research:  
\begin{itemize}
\item We will develop the methods on another CNN framework with GPU computing for larger complex data set.
\item We need to investigate another optimum learning parameters on more complex CNN architecture, i.e.,  dropout and dropconnect regularization,  decay parameters. 
\end{itemize}

\section{Acknowledgment}
This work is supported by Higher Education Center of Excellence Research Grant funded Indonesia Ministry of Research and Higher Education Contract No. 1068/UN2.R12/ HKP.05.00/2016

\section{Conflict of Interests}
The authors declare that there is no conflict of interest regarding the publication of this paper.


\bibliography{library}

\bibliographystyle{abbrv}

\end{document}